# Constructing multi-modality and multi-classifier radiomics predictive models through reliable classifier fusion

Zhiguo Zhou, Zhi-Jie Zhou, Hongxia Hao, Shulong Li, Xi Chen, You Zhang, Michael Folkert, and Jing Wang

**Abstract**— Radiomics aims to extract and analyze large numbers of quantitative features from medical images and is highly promising in staging, diagnosing, and predicting outcomes of cancer treatments. Nevertheless, several challenges need to be addressed to construct an optimal radiomics predictive model. First, the predictive performance of the model may be reduced when features extracted from an individual imaging modality (e.g. PET, CT, MRI) are blindly combined into a single predictive model. Second, because many different types of classifiers are available to construct a predictive model, selecting an "optimal" classifier for a particular application is still challenging. In this work, we developed multi-modality and multi-classifier radiomics predictive models that address the aforementioned issues in currently available models. Specifically, a new reliable classifier fusion strategy was proposed to optimally combine output from different modalities and classifiers. In this strategy, modality-specific classifiers were first trained, and an analytic evidential reasoning (ER) rule was developed to fuse the output score from each modality to construct an optimal predictive model. One public datasets and two clinical case studies were performed to validate model performance. The experimental results indicated that the proposed ER rule based radiomics models outperformed the traditional models that rely on a single classifier or simply use combined features from different modalities.

——————————————— ◆ ———————————————

## 1 INTRODUCTION

With recent developments in medical imaging technology and high-throughput computing, large numbers of quantitative features can be extracted from tomographic images such as PET, CT, and MR images [1] to improve diagnostic accuracy. Radiomics aims to extract and analyze large amounts of information from medical images using advanced quantitative feature analysis [2, 3], and is an emerging field for quantitative cancer screening and diagnosis [4]. Diagnostic accuracy and efficiency for cancer could be improved greatly by using radiomics-driven approaches for a more objective and quantitative assessment. Despite its potential for clinical application, several challenges still need to be addressed to construct an optimal radiomics predictive model.

First, most radiomics-based approaches simply combine the features extracted from different available modalities (such as PET, CT, and MRI) and used them as input for a single classifier. However, different imaging modalities measure different intrinsic characteristics of a lesion. For example, FDG-PET scanning measures glucose metabolism, and CT scanning provides attenuation coefficient information to x-rays. A simple combination of the features extracted from these different modalities may not yield an optimal predictive model. In this work, we developed a modality-specific model first and then obtained reliable fusion results by combining the output of modality-specific models through a recently developed evidential reasoning rule [5].

Second, a single classifier is typically used while constructing a radiomics model. However, many different types of classifiers are available and a "preferred" classifier is often application or problem-specific [6]. Logistic regression [7], random forest, naïve Bayesian, and K-nearest neighbors [8] were used to train predictive models in lung cancer. Twelve different classifiers were used to construct predictive models in head and neck cancer [9]. Therefore, one challenge of current radiomics-based models is how to choose an "optimal" or a "preferred" classifier for a particular application. Instead of trying to select an optimal classifier, we propose a multi-classifier model that maximally utilizes information extracted by different classifiers. According to this strategy, if one classifier is considered as the "expert", a combination of decisions from multiple "experts" will yield a more reliable result.

In both multi-modality and multi-classifier models, information extracted from different sources (e.g., modality

———————————————

- *Zhiguo Zhou, You Zhang, Michael Folkert and Jing Wang are with the Department of Radiation Oncology, UT Southwestern Medical Center, Dallas, TX, USA. E-mail: Zhiguo.Zhou@ UTSouthwestern.edu; You.Zhang@UTSouthwestern.edu; MichaelR.Folkert@UTSouthwestern.edu; Jing.Wang@UTSouthwestern.edu.*
- *Zhi-Jie Zhou is with the High-Tech Institute of Xi'an, Xi'an, China, and also with Department of Information and Control Engineering, Xi'an University of Technology, Xi'an, China. E-mail: zhouzj04@163.com.*
- *Hongxia Hao is with the School of Computer Science and Technology, and with Key Laboratory of Intelligent Perception and Image Understanding of Ministry of Education, Xidian University, Xi'an, China. E-mail: hxhao@xidian.edu.cn.*
- *Shulong Li is with the School of Biomedical Engineering, Southern Medical Unversity, Guangzhou, China. E-mail: Shulong@smu.edu.cn.*
- *Xi Chen is with the Institute of Image Processing and Pattern Recognition, Xi'an Jiaotong Unversity, Xi'an, China. E-mail: xi_chen@mail.xjtu.edu.cn.*

or classifier) need to be combined to yield a final prediction result. Although it was originally designed to combine information from different classifiers, the classifier fusion strategy is an effective solution for both multi-modality and multi-classifier models [10-12]. Currently, classifier fusion strategies often only rely on the weight or relative importance of different classifiers without considering the reliability of the output from individual classifier. In this work, we proposed a new reliable classifier fusion (RCF) strategy that does not only consider the relative importance of different classifiers, but also takes the reliability of output of individual classifier into account so as to get more reliable fusing results. In this strategy, the reliability of the individual classifier output is first defined by considering the output labels from other classifiers. If the output label of one classifier is the same as majority classifiers, the reliability of this classifier should be high. Then both reliability and weight are combined with the output scores from each individual classifier to generate a final output score by an evidential reasoning (ER) rule [5]. Different from the previous developed recursive ER rule [5], which is non-straightforward to train the weight and model, we proposed an analytic ER (AER) rule with an inferenced analytical expression. The proposed reliable classifier fusion strategy was applied to a public UCI dataset [13], a multi-modality model for lung cancer and a multi-classifier model for cervical cancer treatment outcome prediction.

Our major findings are summarized as follows:

1) A new reliable classifier fusion strategy was proposed.

2) An analytic ER rule was inferenced to facilitate model and parameters training.

3) Multi-modality and multi-classifier radiomics predictive models were developed to make radiomics be more reliable.

The remainder of this paper is organized as follows. Work related to the classifier fusion and ER rule are described in section 2. The new RCF strategy is outlined in section 3. Reliability and weight were determined and the analytic ER rule was inferenced in this section. Multi-modality and multi-classifier radiomics predictive models are presented in section 4. The experimental studies on a public dataset study and two real case studies are shown in section 5. The future directions of radiomics and reliable classifier fusion, and conclusions are given in section 6.

## 2 RELATED WORKS

The works related to classifier fusion are described first. As the proposed classifier fusion strategy was proposed according to the ER rule, a brief description is summarized in section 2.2.

### 2.1 Related classifier fusion methods

Based on the type of classifier output, classifier fusion strategies can be divided into three levels, namely abstract, rank, and measurement. Each level requires different kinds of combination rules [14] [15].

Assume there are $K$ individual classifiers and $J$ classes. At the abstract level, the output of the classifier is a unique label corresponding to the class. Each individual classifier is labeled as $j_k$ and the final label is generated by combining all these labels. The typical strategies include the majority vote rule [16] and the Dempster-Shafer (DS) theory-based combination rule [14, 15]. At the rank level, the labels obtained from $K$ individual classifiers are ranked and the top one is considered as the first choice. The classic ranking methods are the Borda count [17] and logistic regression [18]. At the measurement level, the individual classifier assigns a score to each label and combines these scores appropriately. Several methods have been proposed for this level, such as maximum/minimum/sum/product combinations of posterior probabilities [10, 19-21], fuzzy integral for combination [22-24], and DS-based classifier fusion [25-27].

The objective of the three levels of classifier fusion is to obtain the correct label as measured by accuracy. However, most of these methods do not consider the reliability of the individual classifier output or how to obtain more reliable output scores. Thus, a new reliable classifier fusion strategy is proposed to fully consider the reliability of the output scores from individual classifiers.

### 2.2 ER rule

The ER rule is a generalization of the ER algorithm [28, 29] that was originally developed for multiple criteria decision analysis based on the D-S theory [30]. The ER algorithm has been applied successfully in clinical decision support [31, 32], complex system modelling [33-35], and classification [36]. The ER rule is a generic probabilistic reasoning process and can be used to combine multiple pieces of independent evidence with both weight and reliability. In theory, the reliability of each piece of evidence measures its inherent quality of the information source, providing a correct assessment or solution for a given problem. Weight reflects its relative importance compared with other pieces of evidence, while the reliability is the inherent property of the evidence. So far, the ER rule has been applied to multiple group decision analysis [37], safety assessment [38], and data classification [39]. A brief description of the ER rule is shown as follows.

Assume that $\Theta = \{h_1, h_2, \cdots, h_H\}$ is a set of mutually exclusive and collectively exhaustive hypotheses, where $\Theta$ is referred to as a frame of discernment. The power set of $\Theta$ consists of all its subsets, denoted by $P(\Theta)$ or $2^\Theta$, as follows:

$$P(\Theta) = 2^\Theta = \{\emptyset, \{h_1\}, \cdots, \{h_H\}, \{h_1, h_2\}, \cdots, \{h_1, h_M\}, \cdots, \{h_1, h_{H-1}\}, \Theta\}, \quad (1)$$

where $\{h_1, h_2\}, \cdots, \{h_1, h_M\}, \cdots, \{h_1, h_{H-1}\}$ are the local ignorance. In the ER rule, a piece of evidence $e_i$ is represented as a random set and profiled by a belief distribution (BD), as:

$$e_i = \{(\theta, p_{\theta,i}), \forall \theta \subseteq \Theta, \sum_{\theta \subseteq \Theta} p_{\theta,i} = 1\}, \quad (2)$$

where $(\theta, p_{\theta,i})$ is an element of evidence $e_i$, indicating that the evidence points to proposition $\theta$, which can be any subset of $\theta$ or any element of $P(\Theta)$ except from the empty set, to the degree of $p_{\theta,i}$ referred to as probability or degree of belief, in general. $(\theta, p_{\theta,i})$ is referred to as a focal element of $e_i$ if $p_{\theta,i} > 0$.

The reasoning process in the ER rule is performed by defining a weighted belief distribution with reliability (*WBDR*) [40]:

$$m_i = \{(\theta, \tilde{m}_{\theta,i}), \forall \theta \subseteq \Theta; (P(\Theta), \tilde{m}_{P(\Theta),i}\}, \quad (3)$$

where $\tilde{m}_{\theta,i}$ measures the degree of support for $\theta$ from $e_i$ with both weight and reliability being taken into account, defined as follows:

$$\tilde{m}_{\theta,i} = \begin{cases} 0, & \theta = \emptyset \\ c_{rw,i} m_{\theta,i}, & \theta \subseteq \Theta, \theta \neq \emptyset \\ c_{rw,i}(1 - r_i), & \theta = P(\Theta) \end{cases}$$

$$or\ \tilde{m}_{\theta,i} = \begin{cases} 0, & \theta = \emptyset \\ \tilde{w}_i p_{\theta,i}, & \theta \subseteq \Theta, \theta \neq \emptyset \\ 1 - \tilde{w}_i, & \theta = P(\Theta) \end{cases}, \quad (4)$$

$m_{\theta,i} = w_i p_{\theta,i}$ and $c_{rw,i} = 1/(1 + w_i - r_i)$ is a normalization factor, which satisfies $\sum_{\theta \subseteq \Theta} \tilde{m}_{\theta,i} + \tilde{m}_{P(\Theta),i} = 1$. $\tilde{w}_i = c_{rw,i} w_i$ is acting as a new weight. Then the ER rule combines multiple pieces of evidence recursively. If two pieces of evidence $e_1$ and $e_2$ are independent, $e_1$ and $e_2$ jointly support proposition $\theta$ denoted by $p_{\theta,e(2)}$, which is generated as follows:

$$p_{\theta,e(2)} = \begin{cases} 0 & \theta = \emptyset \\ \frac{\hat{m}_{\theta,e(2)}}{\sum_{D \subseteq \Theta} \hat{m}_{D,e(2)}} & \theta \subseteq \Theta, \theta \neq \emptyset \end{cases}, \quad (5)$$

$$\hat{m}_{\theta,e(2)} = [(1 - r_2)m_{\theta,1} + (1 - r_1)m_{\theta,2}] + \sum_{B \cap C = \theta} m_{B,1} m_{C,2}\ \forall \theta \subseteq \Theta, \quad (6)$$

When there are $L$ pieces of independent evidence, the jointly support proposition $\theta$ denoted by $\hat{m}_{\theta,e(L)}$ can be generated by the following two equations:

$$\hat{m}_{\theta,e(L)} = [(1 - r_i)m_{\theta,e(i-1)} + m_{P(\Theta),e(i-1)} m_{\theta,i}] + \sum_{B \cap C = \theta} m_{B,e(i-1)} m_{C,i}, \ \forall \theta \subseteq \Theta, \quad (7)$$

$$\hat{m}_{P(\Theta),e(L)} = (1 - r_i)m_{P(\Theta),e(i-1)}, \quad (8)$$

After obtaining the normalization, the combined $BD\ p_\theta$ can be calculated by the following equation:

$$p_\theta = \frac{\hat{m}_{\theta,e(L)}}{1 - \hat{m}_{P(\Theta),e(L)}}, \ \forall \theta \subseteq \Theta. \quad (9)$$

The recursive formula of the ER rule can combine multiple pieces of evidence in any order [5]. In RCF, each classifier output scores can be considered as one piece of evidence and the final output is obtained by combining all the individual classifier outputs using the ER rule.

## 3 RELIABLE CLASSIFIER FUSION

In RCF, the aim is to obtain a reliable score, which is different from traditional classifier fusion strategies. Therefore, its definition is given as follows:

Definition 1 (Reliable Classifier Fusion). *The RCF fuses the output scores of all individual classifiers with both weight and reliability to obtain more reliable scores.*

In RCF, reliability represents the ability to assess or solve a given problem, while weight is the relative importance to other information sources. An overview of the proposed strategy is shown in Fig. 1. Assume that there are $N$ individual classifiers, and the corresponding output score is $P_j, j = 1, \cdots, N$. For each individual classifier, the relative weight denoted by $w_j, j = 1, \cdots, N$ is obtained by training, while the reliability is denoted by $r_j, j = 1, \cdots, N$. For a test sample, the final score $P_f$ can be obtained using the analytic ER (AER) rule:

$$P_f = AER(P_j, w_j, r_j), j = 1, \cdots, N, \quad (10)$$

The final the label $L_f$ is obtained by:

$$L_f = max P_f. \quad (11)$$

To implement RCF, weight and reliability need to be calculated (defined and determined in Section 3.1). To facilitate model and weight training, the AER rule is inferenced in Section 3.2. In the following subsection, a numerical study is illustrated to show the fusion results. Finally, the sensitivity analysis of reliability is given.

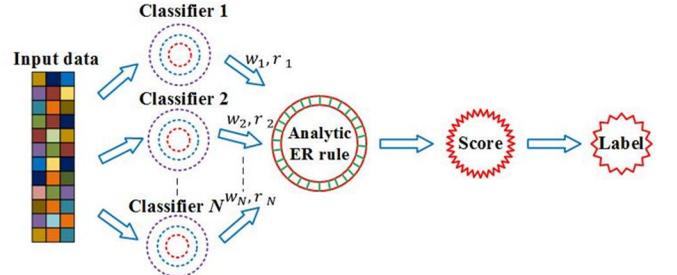

Fig. 1. Flowchart of reliable classifier fusion.

### 3.1 Determination of weight and reliability

According to the description of the ER rule, independent output among individual classifiers is needed. The implication of independence is defined below.

Definition 2 (Independence). *The independence in the RCF indicates that the output by one individual classifier will not be changed and is not related to the output of other classifiers.*

Because the output of individual classifier is independent, the ER rule can be used for fusing. Assume that there are $N$ classifiers denoted by $C = \{C_1, \cdots, C_N\}$. As the predictive outcomes are binary in many radiomics studies, for each individual classifier i, output scores for class I $p_1^i$ and class II $p_2^i$ satisfy $p_1^i + p_2^i = 1, i = 1, \cdots, N$. The output labels are denoted $L = \{l_1, \cdots, l_N\}$. The weight for each classifier denoted by $w_i$ should satisfy the following constraints:

$$0 \leq w_i \leq 1, \ i = 1, \cdots, N. \quad (12)$$

In contrast to most available fusion methods, weights are not necessarily normalized to 1 in RCF. This may increase the search space and improve the model performance during the training process. As weight represents the rel-ative importance of individual classifier, it can be obtained during the model training.

Assuming that the reliability for each classifier is denoted by $r_i, i = 1, \cdots, N$ a definition is given below.

Definition 3 (Reliability). *The reliability of an individual classifier is defined as the similarity between the individual classifier output score and other classifiers output score, which satisfies the following conditions*:

$$r_i = \begin{cases} 0 & \text{when } l_i \neq l_j; j = 1, \cdots, N, j \neq i \\ 1 & \text{when } l_i = l_j \wedge p_{l_j} = 1; j = 1, \cdots, N, j \neq i \\ 0 < r_i < 1 & \text{in other situations} \end{cases}, (13)$$

Based on the above equation, when all the labels of other individual classifiers are different from $l_i$, reliability $r_i$ is defined as 0. When all the labels of other classifiers are the same as $l_i$ and the corresponding output scores are 1, reliability $r_i$ is defined as 1. Other situation is between the above two extreme situations, with the reliability $0 < r_i < 1$.

Definition 3 indicates to what degree an individual classifier is reliable based on the output score of other classifiers. The reliability of the individual classifier depends on whether its output score is close to the output scores of other classifiers. If the output score of one classifier is close to most output scores of other classifiers, then the reliability of this classifier is high. The formula for calculating reliability is given as follows.

We first define the dissimilarity $D_i$ between $C_i$ and $C_j, j = 1, \cdots, N, j \neq i$ as:

$$D_i = \prod_{j=1, j \neq i}^{N}(1 - p_j), i = 1, \cdots, N, \quad (14)$$

where $p_j$ is the output score of classifier $C_i$. Similarity $S_i$ is calculated as:

$$S_i = 1 - D_i = 1 - \left(\prod_{j=1, j \neq i}^{N}(1 - p_j)\right), i = 1, \cdots, N, \quad (15)$$

Assuming that *SL* is the number of classifiers (*SL≤N-1*) that have the same labels with $C_i$, the reliability is calculated as:

$$r_i = \frac{SL}{N-1} \cdot S_i = \frac{SL}{N-1} \cdot \left(1 - \left(\prod_{j=1, j \neq i}^{N}(1 - p_j)\right)\right), i = 1, \cdots, N. \quad (16)$$

The effect of the label information from different classifiers is introduced into the reliability by equation (16). Moreover, proof is given that the reliability calculation formula meets Definition 3.

Theorem 1. *The reliability calculation formula meets Definition 3.*

Proof. *For classifier $C_i$*:

Situation 1: *when all the labels of other classifiers are different from $l_i$, SL=0. According to Eq. (16), it is obvious that $r_i = 0$.*

Situation 2: *when all the labels of other individual classifiers are the same as $l_i$, SL=N-1. Meanwhile, as $p_j = 1, j = 1, \cdots, N, j \neq i$, $S_i = 1$. Therefore, based on Eq. (16), $r_i = 1$.*

Situation 3: *As $0 < p_j < 1$,*

$$0 < \prod_{j=1, j \neq i}^{N}(1 - p_j) < 1, \quad (17)$$

*Then*

$$0 < 1 - \left(\prod_{j=1, j \neq i}^{N}(1 - p_j)\right) < 1, \quad (18)$$

As $0 < SL < N - 1$,

$$0 < \frac{SL}{N-1} < 1, \quad (19)$$

*So*

$$0 < r_i = \frac{SL}{N-1} \cdot \left(1 - \left(\prod_{j=1, j \neq i}^{N}(1 - p_j)\right)\right) < 1,$$
$$i = 1, \cdots, N. \quad (20)$$

After determining the weight and reliability, the AER rule is used to combine the output score.

### 3.2 Analytic ER rule

As there's no local ignorance in outcome prediction, they are pruned in ER rule. Under no local ignorance, the *BD* for each evidence $e_i$ is reduced to the following format:

$$e_i = \left\{(\theta_h, p_{h,i}), h = 1, \cdots, H; \sum_{\theta_h=1}^{M} p_{h,i} = 1\right\}, i = 1, \cdots, N, \quad (21)$$

And *WBDR* is reduced to:

$$m_i = \{(\theta_h, \widetilde{w}_i p_{h,i}), h = 1, \cdots, H; (P_i(\Theta), (1 - \widetilde{w}_i))\},$$
$$i = 1, \cdots, N. \quad (22)$$

where $\theta_h$ is the class and $p_{h,i}$ is the corresponding output score of individual classifier $i$. $\widetilde{w}_i$ is the new weight.

Since normalization in the evidence combination can be applied at the end of the process without changing the combination result, we do not consider normalization when combining all the evidence but apply it in the end. Assume that $\widehat{m}_{\theta_h, l}, h = 1, \cdots, H$ and $\widehat{m}_{P(\Theta), l}$ denote the *WBDR* generated by combining the first $l$ evidence. We first consider a condition of $l = 2$: the combination of two evidences (output scores from two classifiers) without normalization. The combined *WBDR* generated by aggregating the two evidences by orthogonal sum operation are given as follows.

$$\widehat{m}_{\theta_h, 2} = m_{\theta_h, 1} m_{P(\Theta), 2} + m_{\theta_h, 2} m_{P(\Theta), 1} + m_{\theta_h, 1} m_{\theta_h, 2}$$
$$= m_{\theta_h, 1}(m_{\theta_h, 2} + m_{P(\Theta), 2}) + m_{\theta_h, 2} m_{P(\Theta), 1}$$
$$= m_{\theta_h, 1}(m_{\theta_h, 2} + m_{P(\Theta), 2})$$
$$\quad + m_{P(\Theta), 1}(m_{\theta_h, 2} + m_{P(\Theta), 2})$$
$$\quad - m_{P(\Theta), 1} m_{P(\Theta), 2}$$
$$= (m_{\theta_h, 1} + m_{P(\Theta), 1})(m_{\theta_h, 2} + m_{P(\Theta), 2})$$
$$\quad - m_{P(\Theta), 1} m_{P(\Theta), 2}$$
$$= \prod_{i=1}^{2}(m_{\theta_h, i} + m_{P(\Theta), i}) - \prod_{i=1}^{2} m_{P(\Theta), i}, \quad (23)$$

And,

$$\widehat{m}_{P(\Theta), 2} = \prod_{i=1}^{2} m_{P(\Theta), i}, \quad (24)$$

Assume that the following equations are true for the (*l-1*)

evidences. Let $l_1 = l - 1$ and:

$$\hat{m}_{\theta_h,l_1} = \prod_{i=1}^{l-1}(m_{\theta_h,i} + m_{P(\Theta),i}) - \prod_{i=1}^{l-1} m_{P(\Theta),i}, \quad (25)$$

$$\hat{m}_{P(\Theta),l_1} = \prod_{i=1}^{l-1} m_{P(\Theta),i} \quad (26)$$

The above combined probability masses are further aggregated with the *l*th evidence. The combined probability masses are given as:

$$\hat{m}_{\theta,l} = \hat{m}_{\theta_h,l_1} m_{\theta_h,l} + \hat{m}_{\theta_h,l_1} m_{P(\Theta),l} + m_{\theta_h,l} \hat{m}_{P(\Theta),l_1}$$
$$= \hat{m}_{\theta_h,l_1}(m_{\theta_h,l} + m_{P(\Theta),l}) + m_{\theta_h,l} \hat{m}_{P(\Theta),l_1}$$
$$= \hat{m}_{\theta_h,l_1}(m_{\theta_h,l} + m_{P(\Theta),l}) + \hat{m}_{P(\Theta),l_1}(m_{\theta_h,l} + m_{P(\Theta),l}) - \hat{m}_{P(\Theta),l_1} m_{P(\Theta),l}$$
$$= (\hat{m}_{\theta_h,l_1} + \hat{m}_{P(\Theta),l_1})(m_{\theta_h,l} + m_{P(\Theta),l}) - \hat{m}_{P(\Theta),l_1} m_{P(\Theta),l}$$
$$= (m_{\theta_h,l} + m_{P(\Theta),l})(\prod_{i=1}^{l-1}(m_{\theta_h,i} + m_{P(\Theta),i}) - \prod_{i=1}^{l-1} m_{P(\Theta),i} + \prod_{i=1}^{l-1} m_{P(\Theta),i}) - m_{P(\Theta),l} \prod_{i=1}^{l-1} m_{P(\Theta),i}$$
$$= \prod_{i=1}^{l}(m_{\theta_h,i} + m_{P(\Theta),i}) - \prod_{i=1}^{l} m_{P(\Theta),i}, \quad (27)$$

And,
$$\hat{m}_{P(\Theta),l} = m_{P(\Theta),l} \hat{m}_{P(\Theta),l_1}$$
$$= m_{P(\Theta),l} \prod_{i=1}^{l-1} m_{P(\Theta),i} = \prod_{i=1}^{l} m_{P(\Theta),i} \quad (28)$$

Then we normalize the combined *WBDR* results. Assume that $k$ is the normalization factor, therefore:

$$k\left(\sum_{h=1}^{H} \hat{m}_{\theta_h,l} + \hat{m}_{P(\Theta),l}\right) = 1, \quad (29)$$

That is:

$$k\left(\sum_{h=1}^{H}\left(\prod_{i=1}^{l}(m_{\theta_h,i} + m_{P(\Theta),i}) - \prod_{i=1}^{l} m_{P(\Theta),i}\right) + \prod_{i=1}^{l} m_{P(\Theta),i}\right) = 1, \quad (30)$$

$$k\left(\sum_{h=1}^{H}\left(\prod_{i=1}^{l}(m_{\theta_h,i} + m_{P(\Theta),i})\right)\right) - k(H-1)\prod_{i=1}^{l} m_{P(\Theta),i} = 1, \quad (31)$$

So,

$$k = \left(\sum_{h=1}^{H}\left(\prod_{i=1}^{l}(m_{\theta_h,i} + m_{P(\Theta),i})\right) - (H-1)\prod_{i=1}^{l} m_{P(\Theta),i}\right)^{-1}, \quad (32)$$

Therefore,

$$m_{\theta,l} = k\hat{m}_{\theta,l}, \quad m_{P(\Theta),l} = k\hat{m}_{P(\Theta),l}, \quad (33)$$

where $m_{\theta,l}$ and $m_{P(\Theta),l}$ are the combined *WBDR* after normalization. So the *BD* $p_m$ after combining *l* evidence is:

$$p_h = \frac{m_{\theta,l}}{1-m_{P(\Theta),l}} = \frac{k\hat{m}_{\theta,l}}{1-k\hat{m}_{P(\Theta),l}} =$$
$$\frac{k(\prod_{i=1}^{l}(m_{\theta_h,i} + m_{P(\Theta),i}) - \prod_{i=1}^{l} m_{P(\Theta),i})}{1-k\prod_{i=1}^{l} m_{P(\Theta),i}}, h = 1, \cdots, H, \quad (34)$$

Based on Eq. (4) and $l = N$, the final *BD* is:
$$p_h = \frac{k\left[\prod_{i=1}^{N}\left(\frac{w_i p_{h,i}}{1+w_i-r_i} + \frac{1-r_i}{1+w_i-r_i}\right) - \prod_{i=1}^{N}\left(\frac{1-r_i}{1+w_i-r_i}\right)\right]}{1-k\prod_{i=1}^{N}\left(\frac{1-r_i}{1+w_i-r_i}\right)}, h = 1, \cdots, H, \quad (35)$$

and $k$ is:

$$k = \left[\sum_{h=1}^{H}\left(\prod_{i=1}^{N}\left(\frac{w_i p_{h,i}}{1+w_i-r_i} + \frac{1-r_i}{1+w_i-r_i}\right)\right) - (H-1)\prod_{i=1}^{N}\left(\frac{1-r_i}{1+w_i-r_i}\right)\right]^{-1}. \quad (36)$$

Eqs. (35) and (36) represent the AER rule without local ignorance, providing an explicit aggregation function. The AER rule can be used in many situations such as training model and weight.

### 3.3 Numerical comparison study

In this subsection, a simple example illustrates the discrimination of the RCF. The classic weighted fusion model (named as WF) is also shown for comparison, which is expressed as:

$$P_f = \sum_{j=1}^{N} P_j w_j, \quad (37)$$

where $P_j$ is the individual classifier output scores and $w_j$ is the relative weight. Assume that there are three individual classifiers with two group output scores obtained by 2-cross-validation as shown in Table 1.

TABLE 1
TWO GROUP OUTPUT SCORES FOR THREE INDIVIDUAL CLASSIFIERS BY 2-CROSS-VALIDATION

|              | Group 1 | | Group 2 | |
| --- | --- | --- | --- | --- |
|              | Class I | Class II | Class I | Class II |
| Classifier 1 | 0.8 | 0.2 | 0.8 | 0.2 |
| Classifier 2 | 0.7 | 0.3 | 0.3 | 0.7 |
| Classifier 3 | 0.6 | 0.4 | 0.4 | 0.6 |

At first, assuming that the weights for three individual classifiers are the same and normalized, $\mathbf{w} = (\frac{1}{3}, \frac{1}{3}, \frac{1}{3})$. Based on the developed reliability formula, reliability for group 1 is {0.88, 0.92, 0.94}, while reliability for group 2 is {0, 0.34, 0.38}. As the labels of the other two classifiers are different from classifier 1 in group 2, the reliability of classifier I is 0. Meanwhile, we set reliability to 1 for all the individual classifiers (named as RCF-1) in RCF to show how reliability effects discrimination. The fusing output for the three strategies is shown in table 2. In group 1, output labels are same as the three individual classifiers, and the output score for the same label in RCF is higher than WF and RCF-1, indicating that RF is more discriminatory. For group 2, because the labels are the same in two individual classifiers, the fusing results are more reasonable in RCF. Most current classifier fusion methods without reliability can be considered that the reliability of all the individual classifiers is 1. In fact, as reliability represents the ability to provide the correct assessment or solution, it should be different for each individual classifier output.

TABLE 2
FUSING RESULTS OF THREE INDIVIDUAL CLASSIFIERS FOR THREE STRATEGIES

|  | Group 1 | | Group 2 | |
| --- | --- | --- | --- | --- |
|  | Class I | Class II | Class I | Class II |
| RCF | 0.8393 | 0.1607 | 0.4592 | 0.5408 |
| WF | 0.7 | 0.3 | 0.5 | 0.5 |
| RCF-1 | 0.7778 | 0.2222 | 0.5333 | 0.4667 |

### 3.4 Sensitivity analysis of reliability

As mentioned earlier, reliability contributes to output scores. Ideally, the higher the reliability, the more discriminatory are the output scores. For this reason, a sensitivity analysis shows how final output scores are influenced by changing reliability.

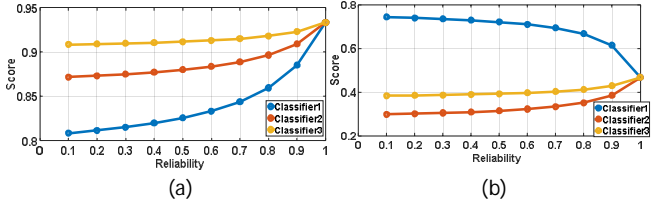

Fig. 2. Sensitivity analysis of reliability: (a) Group 1; (b) Group 2.

To facilitate the sensitivity analysis, we changed reliability from 0.1 to 1 with 0.1 as a step (shown in Table 1). To clearly show the sensitivity, we only changed the reliability of one classifier, while another two were set as 1. The weight for three classifiers is still the same. The sensitivity analysis of the reliability is shown in Fig. 2. Group 1 shows the output of class I with the reliability changes for three classifiers, while group 2 shows the output of class II. As the labels in group 1 are same in the three classifiers, the output score will increase with the increasing of reliability. In group 2, the output of classifier 1 decreases when reliability increases. This is because the label of classifier 1 is different from the other two classifiers; an increase in reliability of classifier 1 decreases the output score for class II. Overall, reliability is important for the output score, making the fusion results more reliable and discriminatory.

## 4 PREDICTIVE MODEL CONSTRUCTION

In this section, we constructed multi-modality and multi-classifier radiomics predictive models. In these models, RCF is the key component for integrating the outputs from different modalities or classifiers. Meanwhile, other procedures such as feature extraction and selection are also needed and will be described in the following subsections.

### 4.1 Multi-modality predictive model

The training process of the multi-modality model mainly consists of three stages, namely feature extraction, feature selection, and predictive model construction (Fig. 3). Assume that there are three modalities, including two image modalities such as PET and CT, and clinical parameters.

Stage 1: Extracting features from images.

The imaging features, including intensity, texture, and geometric features [41], are extracted from segmented tumors, and the extracted features for the three modalities are denoted by $fea^i = \{fea_1^i, fea_2^i, \cdots, fea_{M_i}^i\}, i = 1,2,3$.

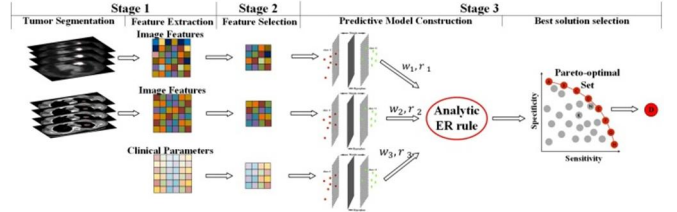

Fig. 3. Multi-modality model training stage.

Stage 2: Selecting features for each modality.

To select the features from $fea^i, i = 1,2,3$ that can achieve optimal performance for each classifier, the multi-objective based feature selection method was used. For this model, sensitivity and specificity are considered as objectives simultaneously. Assume that the sensitivity and specificity are denoted by $f_{sen}, f_{spe}$, respectively:

$$f_{sen} = \frac{TP}{TP+FN}, \quad (38)$$

$$f_{spe} = \frac{TN}{TN+FP}, \quad (39)$$

where *TP* is the number of true positives, *TN* is the number of true negatives, *FP* is the number of false positives, and *FN* is the number of false negatives [42]. SVM is used to construct the predictive model, and the goal is to simultaneously maximize $f_{sen}^i$ and $f_{spe}^i$ to obtain the Pareto-optimal solution set:

$$f_{fea}^i = \max_{fea^i}(f_{sen}^i, f_{spe}^i), i = 1,2,3. \quad (40)$$

To solve the problem above, we used an iterative multi-objective immune algorithm (IMIA) [41, 43]. IMIA is a type of multi-objective evolutionary algorithm that has shown superior performance for multi-objective optimization [44] in recent years. The IMIA procedure is summarized below and detailed implementation has been described previously [41].

1). Initialization. The initial solution set was generated randomly. Each solution consists of a group of binary values named as "individual". In each individual, "1" indicates that the feature is selected, while "0" indicates that the feature is not selected. Assume that the solution set size is **P** and the maximal generation is $G$.

2). Clonal operation. Copy the solution with a larger crowding-distance several times to keep the best solutions.

3). Mutation operation. Perform the static mutation operation to generate the better solutions [32, 45].

4). Deleting operation. Because a newly generated solution set may include the same solutions (same sensitivity and specificity), the unique one should be kept to avoid a search space reduction.

5). Updating the solution set. Select the **P** solutions to keep the size of the solution set, using the AUC based fast nondominated sorting approach.
6). Termination. The algorithm ends once it reaches the maximal generation $G$; Otherwise, go to step 2).

Then, the optimal solution is selected from the Pareto-optimal solution set according to sensitivity, specificity, and AUC; the selected features are also determined. Moreover, to select the most reliable features, we considered the most frequently selected features by running them multiple times as the finally selected ones; the feature number is determined by the average number of selected features in multiple running times. Assume that the number of selected features is $MS_i, i = 1,2,3$, and the selected features are denoted by $fea\_s^i = \{fea\_s_1^i, fea\_s_2^i, \cdots, fea\_s_{M_i}^i\}, i = 1,2,3$:

$$fea\_s^i \subseteq fea^i, i = 1,2,3, \quad (41)$$

Stage 3: Constructing the predictive model.

After selecting the features for each modality, the predictive model and corresponding weights are trained. Assume that the model parameters for three modalities are denoted by $p = (p_1, p_2, \cdots, p_N)$. When the selected features $fea\_s^i, i = 1,2,3$ input the predictive model for each modality, we can obtain the score which is denoted by $S_i = (s_i^1, s_i^2), i = 1,2,3$. Then, reliability $r_i, i = 1,2,3$ is obtained using the reliability calculation method. Assume that the weight is $w_i, i = 1,2,3$. The final score $S = (s^1, s^2)$ is calculated using the AER rule and the label can also be obtained:

$$S = AER(S_i, w_i, r_i), i = 1,2,3, \quad (42)$$

To obtain the optimal model, the parameters and weights mentioned earlier need to be trained. This method is same as the feature selection process. Assume that $f_{sen}^M, f_{spe}^M$ represent the sensitivity and specificity obtained by the output labels and the goal is to maximize the following function:

$$f^M = \max_{p,w}(f_{sen}^M, f_{spe}^M). \quad (43)$$

The test stage consisting of two phases is shown in Fig. 4. In the first phase, the selected features are extracted from the segmented tumor and clinical parameters. The final outcome is predicted after combining the outputs from three modalities using the AER rule in the second phase.

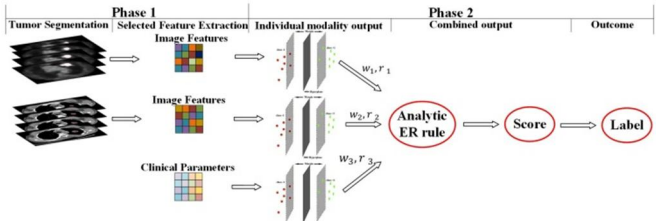

Fig. 4. Multi-modality model test stage.

## 4.2 Multi-classifier predictive model

The training stage of combining different classifiers to construct a predictive model is shown in Fig. 5. As in the case of the multi-modality model, the multi-classifier model also consists of three stages: feature extraction, feature selection, and predictive model construction. The multi-objective based feature selection and predictive model construction were also adopted. Because of its similarities with the multi-modality model, we will mainly describe the differences between the two.

(1). The features are selected for each classifier in stage 2, indicating that the predictive model is constructed by different classifiers such as support vector machine (SVM) and logistic regression (LR). Assuming that the number of individual classifier is $N$, the objective function is:

$$f_{fea}^i = \max_{fea^i}(f_{sen}^i, f_{spe}^i), i = 1, \cdots, N. \quad (44)$$

(2). When constructing the predictive model, the individual classifier is different.

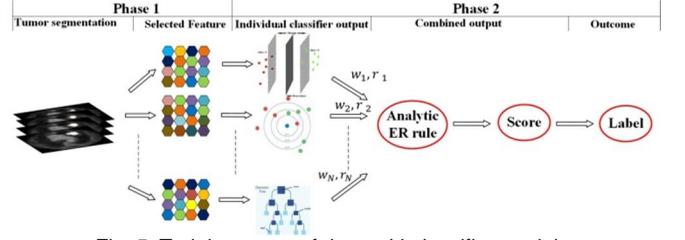

Fig. 5. Training stage of the multi-classifier model.

The test stage is similar with multi-modality test stage (Fig. 6).

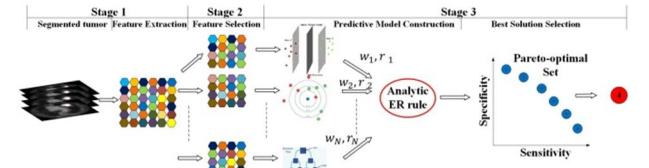

Fig. 5. Multi-classifier model during the test stage.

## 5 EXPERIMENTS

To evaluate the proposed strategy, we conducted two separate experiments. First, we tested five UCI public datasets for binary classification to show the advantage of RCF. Second, we evaluated two case studies to compare the performance of RCF with other well-known strategies in real situations: 1) multi-modality radiomics model for predicting distant failure in lung cancer after stereotactic body radiation (SBRT); and 2) multi-classifier radiomics model for predicting distant failure in cervical cancer after radiotherapy.

### 5.1 UCI dataset study

#### 5.1.1 Experimental setup

In the first study, 5 UCI [13] binary classification datasets were used to validate the performance of RCF (Table 3). In this table, we report the number of instances and features, and also the number of two classes. Several related fusion methods were also evaluated, including the classical WF. Besides, as the ER rule was developed based on the ER [46-48] and DS theories, the ER fusion (ERF) and DS fusion (DSF) [49] were also evaluated for compari comparisons. Six different classifiers including the linear SVM (LSVM), logistic regression (LR), discriminant analysis (DA), decision tree (DT), K-nearest-neighbor (KNN), and naive Bayesian (NB) were taken as multiple classifiers for fusing. In this experiment, 5-cross-validation was performed. Within each training set, 70% of samples were used to train the weight of different classifiers, while the remaining samples were used to validate the system. As the clonal selection algorithm [31] achieves optimal or sub-optimal results, we used it for training with the objective function to maximize the AUC. AUC, sensitivity and specificity were used to evaluate performance. The pro-gram was performed 10 times, during which the mean and standard deviation were calculated for each evaluation criteria.

TABLE 3
SUMMARY OF FIVE UCI DATASETS

| Dataset | Instance number | Feature number | Class1 | Class 2 |
|---|---|---|---|---|
| Heart | 270 | 13 | 150 | 120 |
| Ionosphere | 351 | 34 | 225 | 126 |
| Mask | 476 | 166 | 269 | 207 |
| Sonar | 208 | 60 | 97 | 111 |
| Spambase | 4601 | 57 | 1813 | 2788 |

### 5.1.2 Results and analysis

The predictive results for five datasets with four strategies are presented in table 4, with the best results marked in bold. In all cases, RCF performs better than the four strategies in terms of AUC, indicating that RCF can acquire more reliable and discriminatory results. DSF performs better than the other methods in the Ionosphere and Mask datasets for sensitivity, while RCF is the best in the remaining datasets. RCF also shows the best specificity except for the Sonar dataset. RCF performs better than DSF, and DSF performs better than ERF for the three related strategies. This is because ERF introduces weight and avoids evidence conflict compared to DSF. Beyond this, RCF not only considers the weight, but also introduces reliability, which fully considers the inner and outer attributes of the individual classifier output. As reported earlier [5], both DSF and ERF are special cases of RCF. When the evidence is fully reliable and equal to 1, the RCF is the same as DSF, and RCF is same as ERF when the weight is equal to reliability. Moreover, as the ER rule can handle the evidence conflict adequately, RCF improves performance. The comparative output scores between RCF and the other three fusion strategies of the same class in Heart and Mask datasets are shown in Fig. 7; 200 test samples were chosen for each dataset. In most cases, RCF always obtained higher scores for the positive class and lower scores for the negative class. The more discriminatory scores obtained by RCF are attributed to the reliability proposed in this work.

TABLE 4
PERFORMANCE EVALUATION FOR FOUR FUSION STRATEGIES IN FIVE DATASETS

| Dataset | Strategy | AUC | Sensitivity | Specificity |
|---|---|---|---|---|
| Heart | WF | 0.85±0.02 | 0.70±0.02 | 0.88±0.02 |
|  | DSF | 0.86±0.01 | **0.77±0.02** | 0.87±0.01 |
|  | ERF | 0.86±0.01 | 0.76±0.02 | 0.87±0.01 |
|  | RCF | **0.88±0.01** | **0.77±0.02** | **0.89±0.01** |
| Ionosphere | WF | 0.94±0.02 | 0.78±0.02 | 0.97±0.01 |
|  | DSF | 0.92±0.02 | **0.83±0.02** | 0.94±0.01 |
|  | ERF | 0.95±0.01 | 0.81±0.01 | 0.96±0.01 |
|  | RCF | **0.96±0.01** | 0.82±0.02 | **0.98±0.01** |
| Mask | WF | 0.88±0.02 | 0.76±0.02 | 0.84±0.02 |
|  | DSF | 0.86±0.02 | **0.88±0.02** | 0.68±0.02 |
|  | ERF | 0.91±0.01 | 0.87±0.02 | 0.83±0.02 |
|  | RCF | **0.93±0.01** | 0.86±0.02 | **0.86±0.02** |
| Sonar | WF | 0.8±0.02 | 0.71±0.03 | **0.74±0.03** |
|  | DSF | 0.78±0.02 | 0.78±0.03 | 0.67±0.03 |
|  | ERF | 0.83±0.02 | 0.83±0.02 | 0.69±0.03 |
|  | RCF | **0.85±0.01** | **0.84±0.02** | 0.72±0.02 |
| Spambase | WF | 0.94±0.02 | 0.86±0.03 | **0.92±0.01** |
|  | DSF | 0.94±0.01 | 0.86±0.01 | 0.91±0.00 |
|  | ERF | 0.97±0.00 | 0.93±0.01 | **0.92±0.01** |
|  | RCF | **0.98±0.00** | **0.94±0.01** | **0.92±0.01** |

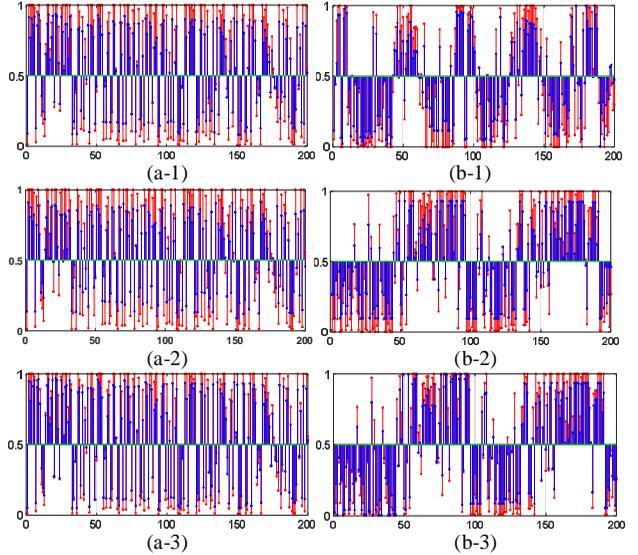

Fig. 7. Comparing output scores between RCF and WF, DSF, ERF for two datasets. The red lines depict RCF, while blue lines indicate another three strategies. Comparing results with WF, DSF (a-1, 2, 3), and ERF for Heart; Mask (b-1, 2, 3).

### 5.2 Case study 1

In this case study, the multi-modality predictive model was used to predict distant failure for lung stereotactic body radiation therapy (SBRT) in early stage non-small cell lung cancer (NSCLC). The problem description and experimental setup are given in the first section, while the experimental results and analysis are given in the second part.

### 5.2.1 Problem description and setup

In recent years, dose escalation with SBRT has become the standard of care for inoperable early stage NSCLC. However, distant failure in early stage patients is still common in about 30% patients, with more than 70% of cases occurring within the first 2 years [50]. Distant failure is a critical oncologic event because it correlates closely with mortality. If the patient is at high risk for early distant failure, additional systemic therapy after SBRT may reduce this risk and improve overall survival. However, therapy-related toxicity may increase mortality since this population usually presents preexisting health conditions. Therefore predicting high risk for early-stage distant failure is essential for this group of patients.

The study involved 52 early IA and IB stage patients who had received SBRT from 2006 to 2012. The follow-up range was from 6 to 64 months, with a median of 18 months. Among these patients, 23.1% had experienced distant failure. The clinical parameters extracted from the clinical charts included (a) demographic parameters, such age, ethnicity, and gender; (b) tumor characteristics, including primary diagnosis, central or peripheral tumor, tumor size, histology, location, and stage; (c) treatment parameters, including number fractions, dose per fraction, and biological equivalent dose (BED); and (d) pre-treatment medications, including anti-inflammatories, antidiabetic, metformin, statin, ACE (Angiotensin-converting-enzyme) inhibitor, and ASA (acetylsalicylic acid).

Tumors are segmented before the image features are extracted. As described previously [41], the middle slice was segmented using the object information-based interactive segmentation method (OIIS) [42] in the first step, and the other slices were segmented by the well-known OTSU method [51] in the second step. Examples of tumor extraction in PET and CT images are illustrated in Fig. 8. The first and third rows indicate the 3D visualization segmentation results, while the second and fourth rows show the corresponding segmentation results in one 2D slice. Three types of imaging features were extracted as follows: (1). Intensity features, including minimum, maximum, mean, stand deviation, sum, median, skewness, kurtosis, and variance; (2). Texture features, including energy, entropy, correlation, contrast, texture variance, sum-mean, inertia, cluster shade, cluster prominence, homogeneity, max-probability, and inverse variance; (3). Geometry features, including volume, major diameter, minor diameter, eccentricity, elongation, orientation, bound box volume, and perimeter. The details of these features were described previously [41]. SVM with radial basis function kernel (RBF) was used to construct the predictive model for each modality.

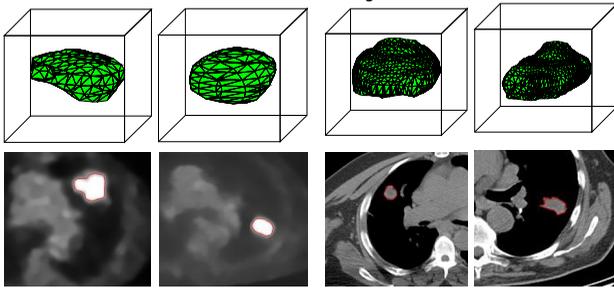

Fig. 8. Extracted tumors from PET and CT images. The first row show 3D PET and CT visualization segmentation results, while the second row show the corresponding results in a 2D slice.

We used the traditional model construction method, which combines the features from all the modalities, to construct the one model (named as single) for comparison. As in the case of the first study, three other strategies were also used to fuse the output of each modality. The parameters of IMIA to select the features and construct the model were the same. They included a population number of 100 and a maximal generation number of 200. In the mutation operator, the mutation probability was set to 0.9. The five-folder cross-validation was performed and the final features were selected by performing them 20 times in the feature selection stage. The model construction stage was performed 10 times, including calculating the mean and standard deviation for each evaluation criteria.

### 5.2.2 Results and analysis

The selected features for the three modalities are provided in table 5. Five features were selected in the clinical parameters, while 15 and 9 features were selected in PET and CT images, respectively. PET features were selected mostly among three modalities, and variance, skewness, correlation, and eccentricity were selected for both PET and CT images.

Examples of the Pareto-optimal set for the three models during the training stage are shown in Fig. 9. The final selected solution is marked in red, while the unselected labels are marked in blue. The ten times performing results of mean and standard deviation values for five models in the testing stage are reported in table 6. Four models combined the fusion strategies and one model combined all the features (named as Single). Our results show that performance is improved when multiple modalities are modelled individually. RCF performed the best among all four fusion strategies.

TABLE 5
SELECTED FEATURES FOR THREE MODALITIES

| Modality | Selected features |
|---|---|
| Clinical parameters | Ethnicity, Gender, Primary Diagnosis, Antiinflammatories, ASA |
| PET features | Maximum, Minimum, Mean, Median, Variance, Skewness, Correlation, Inertia, Cluster shade, Volume, Major diameter, Minor diameter, Eccentricity, Bounding-Box, Volume, Perimeter |
| CT features | Skewness, Kurtosis, Energy, Correlation, Variance, Max-Probability, Eccentricity, Elongation, Orientation |

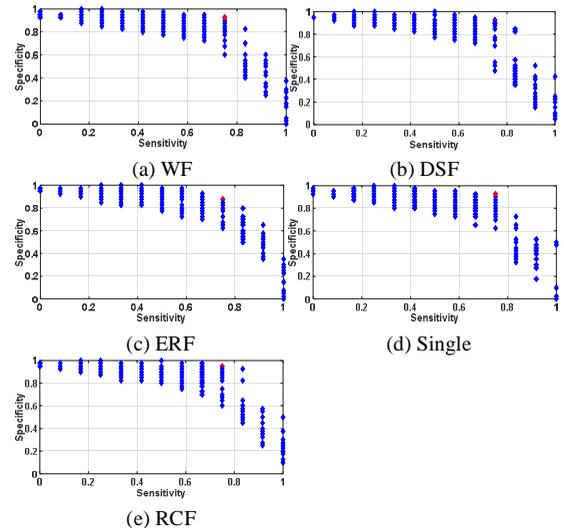

Fig. 9 Pareto-optimal solution set and best solution selection for three methods. Red label: final selected solution; Blue labels: unselected solutions.

TABLE 6
PERFORMANCE EVALUATION FOR FIVE STRATEGIES

|  | AUC | Sensitivity | Specificity |
|---|---|---|---|
| WF | 0.82±0.01 | **0.75±0.00** | 0.91±0.02 |
| DSF | 0.83±0.02 | **0.75±0.00** | 0.91±0.03 |
| ERF | 0.82±0.01 | **0.75±0.00** | 0.91±0.03 |
| Single | 0.82±0.02 | **0.75±0.00** | 0.80±0.04 |
| RCF | **0.86±0.01** | **0.75±0.00** | **0.95±0.01** |

## 5.3 Case study 2

In this study, distant failure for cervical cancer was predicted by the multi-classifier radiomics model. The experimental setup is provided in the first section, and the experimental results and analysis are described in the second part.

### 5.3.1 Experimental setup

Similar to lung SBRT, distant failure is also critical for cervical cancer. The combination of external beam radia-tion therapy (EBRT) and brachytherapy allows maximal dose delivery. However, patients may benefit from supplementing radiation therapy with additional chemotherapy hoping to achieve more durable treatment response and avoid distant failure. In contrast, at least 20% of patients with locally advanced cervical cancer in the pelvis still experience distant failure [52, 53].

This study included 75 patients treated for cervical cancer with definitive intent between 2009 and 2012. Criteria included stage IB1 to IVA disease treated with EBRT or combined with high dose rate (HDR) intracavitary brachytherapy and retrievable pre-treatment PET/CT scanning. In this study, PET imaging was used to predict distant failure in cervical cancer. All tumors were contoured manually by the radiation oncologists. All imaging features were calculated based on SUV. As in the case of study 1, intensity, texture, and geometry features were extracted.

Six individual classifiers, including linear SVM (LSVM), logistic regression (LR), discriminant analysis (DA), decision tree (DT), K-nearest-neighbor (KNN), and naive Bayesian (NB) were used. Different from case study 1, LSVM is used. This is because LSVM doesn't need to train the model parameters and is more convenient to train the model. In KNN, the neighbor number was set to 5. In addition to the three fusion strategies, six individual classifiers were also used for comparison. Parameter setting in IMIA and feature selection was the same as those in case study 1.

### *5.3.2 Results and analysis*

Of the features selected for the six classifiers, DT includes the highest number of features and LR includes the lowest (Table 7). The features most frequently selected among the six classifiers, are maximum, standard deviation, sum-mean, variance, and volume. The Pareto-optimal set and best solutions in four fusion strategies and six individual models are shown in Fig. 10. The experimental results are reported in Table 8. Fusion strategies always perform better than individual classifiers, demonstrating the advantage of combining classifiers. Among the fusion strategies, RCF always obtains the better performance, and AUC shows that RCF is more reliable.

TABLE 7
SELECTED FEATURES FOR SIX INDIVIDUAL CLASSIFIERS

| Individual classifier | Selected features |
|---|---|
| LSVM | Skewness, Energy, Entropy, Variance, Mean, Inertia, Cluster Shade, Cluster, Max-Probability, Inverse Variance, Eccentricity, Elongation, Orientation |
| LR | Maximum, Sum, Contrast, Sum-Mean, Volume |
| KNN | Maximum, Stand deviation, Variance, Kurtosis, Energy, Entropy, Sum-Mean, Homogeneity |
| DA | Maximum, Stand deviation, Kurtosis, Contrast, Sum-Mean, Inertia, volume |
| DT | Maximum, Minimum, Median, Stand deviation, Variance, Sum, Kurtosis, Sum-Mean, Inertia, Cluster Shade, Inverse Variance, Volume, Major diameter, Orientation |
| NB | Maximum, Stand deviation, Variance, Sum-Mean, Cluster, Homogeneity, Volume |

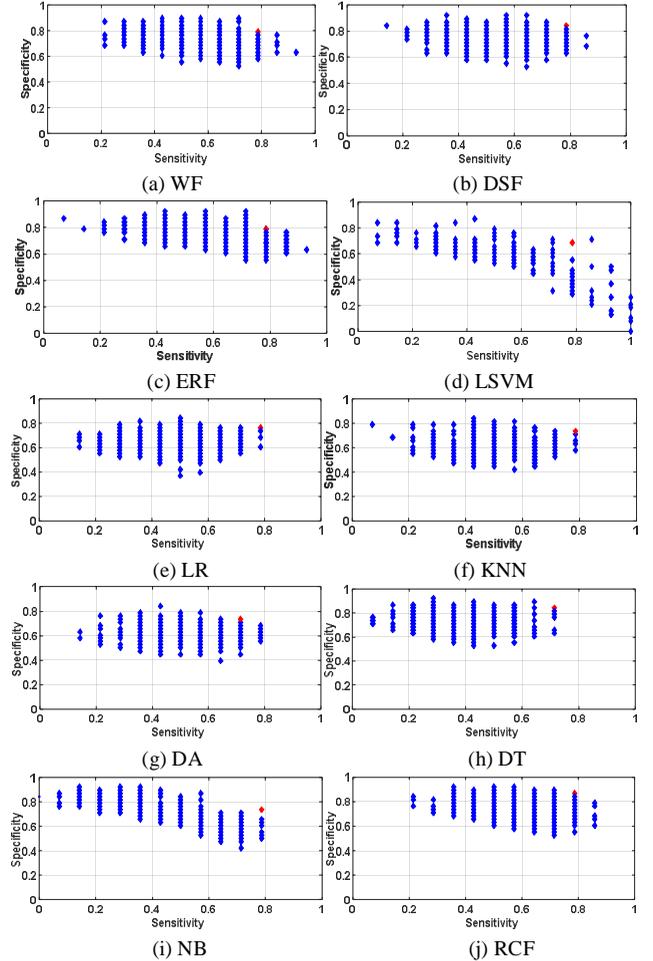

Fig. 10. Pareto-optimal solution set and best solution selection for four fusion strategies and six individual classifiers. Red label: final selected solution; Blue labels: unselected solutions.

TABLE 8
PREDICTIVE RESULTS FOR FUSION STRATEGIES AND INDIVIDUAL CLASSIFIERS

|  | AUC | Sensitivity | Specificity |
|---|---|---|---|
| WF | 0.81±0.02 | **0.79±0.00** | 0.79±0.01 |
| DSF | 0.81±0.02 | **0.79±0.00** | 0.83±0.02 |
| ERF | 0.80±0.02 | **0.79±0.00** | 0.79±0.03 |
| LSVM | 0.73±0.04 | 0.76±0.08 | 0.68±0.05 |
| LR | 0.74±0.03 | 0.74±0.03 | 0.75±0.03 |
| KNN | 0.75±0.04 | 0.78±0.07 | 0.75±0.04 |
| DA | 0.74±0.02 | 0.74±0.03 | 0.74±0.04 |
| DT | 0.76±0.05 | 0.72±0.04 | 0.80±0.04 |
| NB | 0.72±0.03 | 0.76±0.06 | 0.73±0.04 |
| RCF | **0.83±0.02** | **0.79±0.00** | **0.84±0.03** |

## 6 DISCUSSIONS AND CONCLUSION

We constructed multi-modality and multi-classifier predictive models to increase the reliability of radiomics. In these models, different modalities or classifiers were constructed individually and their outputs were combined by the proposed RCF strategy to obtain more discriminatory and reliable output scores. In RCF, reliability was introduced and the analytic ER rule was inferenced to facilitate model training by combining weight, reliability, and output scores. The experiments on UCI public datasets demonstrated that RCF was more reliable and discriminatory than traditional fusion strategies. Two case studies on lung and cervical can-cer patients showed that the proposed models were not only more effective than current radiomics predictive models, but also outperformed other related fusion strategies.

As different classifiers contribute differently to the final fusion result, the weight or relative importance of the different classifiers is considered in most traditional fusion strategies. On the other hand, the inherent properties of an individual classifier also influence the fusion results. Because this aspect is rarely considered in current fusion strategies, we fully considered the inherent property of an individual classifier (defined as reliability) and the relative importance of different classifiers for more reliable and discriminatory fusion results. Hopefully, this preliminary study will stimulate more advanced fusion strategies that consider both reliability and weight. In the proposed reliable fusion strategy, the meaning of reliability may differ depending on the application. Because the calculation formula may also be different, a more suitable reliability calculation method may need to be developed for different applications. Additionally, the presented RCF strategy is used only for binary classification, and more models should be developed for multi-class classification problems.

To integrate the outputs from different sources, the conventional recursive ER rule can also be used to combine various independent pieces of evidence. However, when a model needs training by optimization, the recursive ER rule requires training of the additional parameters, which complicates the process. To address this this, we developed an analytic ER rule. The AER rule delivers more flexibility to the ER rule by aggregating a large number of individual classifiers. Additionally, the AER rule provides a straightforward way to conduct a sensitivity analysis for parameters such as weight and reliability.

In this study, two radiomics predictive models, including multi-modality and multi-classifier models were constructed. Both models were constructed in the way of "multi-". For other applications, such as predicting histological types of lung cancer, multi-class models are needed. In addition to predictive models, other procedures [2, 3] such as image acquisition and storage, tumor segmentation, feature extraction, and selection in radiomics also face "multi-" challenges. Because images are always acquired from multiple devices in terms of acquisition and storage, uncertainties in extracted features may be observed. During feature selection, sensitivity and specificity need to be considered as they play a more important role for performance evaluation in medical applications than accuracy or AUC. Because of the similarity in the "multi-" problems described earlier, a unified "multifaceted radiomics" model is needed to fully address the challenges of radiomics toward practical applications.

## ACKNOWLEDGMENT

This work was supported in part by the American Cancer Society (ACS-IRG-02-196) and the US National Institutes of Health (5P30CA142543). The authors would like to thank Dr. Damiana Chiavolini for editing the manuscript.

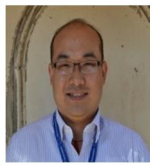
**Zhiguo Zhou** received the B.S. and Ph.D. degrees in Computer Science and Technology from Xidian University, Xi'an, China, in 2008 and 2014, respectively.

He joined the University of Texas Southwestern Medical Center, Dallas, TX, in 2014 as a Postdoctoral Researcher, where he is currently an instructor in Department of Radiation Oncology. He was a visiting scholar with Leiden University, Leiden, the Netherlands, from 2013 to 2014. He has published 16 peer reviewed journal papers. His current research interests include outcome prediction in cancer therapy, radiomics, medical image analysis, deep learning, machine learning and artificial intelligence.

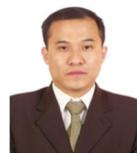
**Zhi-Jie Zhou** received the B.Eng. and M.Eng. degrees from the High-Tech Institute of Xi'an, Xi'an, China, in 2001 and 2004, respectively, and the Ph.D. degree from Tsinghua University, Beijing, China, in 2010, all in control science and management.

He is currently an Associate Professor with the High-Tech Institute of Xi'an. In 2009, he was a Visiting Scholar with the University of Manchester, Manchester, U.K., for six months. He has published approximately 70 articles. His current research interests include belief rule base, dynamic system modeling, hybrid quantitative and qualitative decision modeling, and fault prognosis and optimal maintenance of dynamic systems.

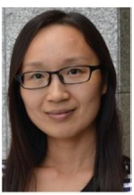
**Hongxia Hao** received the Ph.D. degree from Xidian University, Xian, China, in 2014.

She is currently working in computer application technology at the School of Computer Science and Technology, Xidian University and the Key Laboratory of Intelligent Perception and Image Understanding of Ministry of Education of China, Xi'an, China. Her current interests include machine learning, radiomics and intelligent image processing.

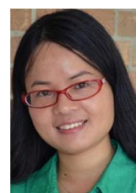
**Shulong Li** received the B. S. in Mathematics from Minnan Normal University, Zhangzhou, China, in 2003 and Ph. D degrees in Mathematics from Sun Yat-sen University, Guangzhou, China, 2008.

She joined the Southern Medical University, Guangzhou, in 2008 as a lecture, where she is currently an associate professor in School of Biomedical Engineering. She is a visiting assistant professor with University of Texas Southwestern Medical Center, Dallas, TX, from 2016 to 2017. She has published more than ten journal papers. Her current research interests include outcome prediction in cancer therapy, radiomics, medical image analysis, deep learning, machine learning and artificial intelligence.

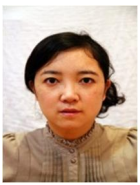
**Xi Chen** received the PhD degree in information and communication engineering in July 2010 from Xi'an Jiaotong University, Xi'an, China.

She is currently assistant professor in Institute of Image Processing & Pattern Recognition, Xi'an Jiaotong University, Xi' an, China. Her research interests include breast imaging, cone-beam CT imaging, radiomics and machine learning.

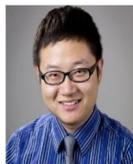
**You Zhang** received his B.S. degree in Physics from Nanjing University, Nanjing, China in 2010 and Ph.D degree in Medical Physics from Duke University, Durham, USA in 2015.

He joined the University of Texas Southwestern Medical Center, Dallas, TX, in 2015 as a medical physics resident in the Department of Radiation Oncology. He has published 19 peer-reviewed journal papers. His current research interests include radiomics; medical image reconstruction, analysis and segmentation; deep learning; and radiotherapy treatment plan optimization.

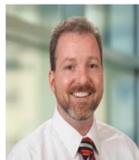
**Michael Folkert** received his Ph.D. in Radiological Sciences from the Massachusetts Institute of Technology in Cambridge, MA USA in 2005 and M.D. from Harvard Medical School in 2009.

He joined the University of Texas Southwestern Medical Center, Dallas, TX, in 2014 as an Assistant Professor in the Department of Radiation Oncology, and serves as the Medical Residency Director, Director of the Intraoperative Radiation Therapy program, and Co-Director of the Brachytherapy program. He has published 37 journal papers and 6 book chapters. His current research interests include outcome prediction in cancer therapy, radiomics/medical image analysis, brachytherapy applicator development, and interventional techniques for toxicity reduction. His clinical interests include management of ocular, gastrointestinal, genitourinary, musculoskeletal, and spine malignancies.

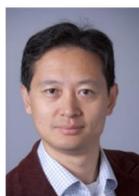
**Jing Wang** received his B.S. degree in Materials Physics from University of Science and Technology of China in 2001, M.A. and Ph.D. degrees in physics from the State University of New York at Stony Brook in 2003 and 2006, respectively. He finished his postdoctoral training in the Department of Radiation Oncology at Stanford University in 2009.

He is currently an Associate Professor and Medical Physicist in the Department of Radiation Oncology at the University of Texas Southwestern Medical Center. Dr. Wang has published more than 60 peer reviewed journal papers. His research focuses on medical imaging and its applications in radiation therapy.